\pdfoutput=1
\documentclass[11pt]{article}
\usepackage[final]{acl}
\usepackage{times}
\usepackage{latexsym}
\usepackage[T1]{fontenc}
\usepackage{float}
\usepackage[utf8]{inputenc}
\usepackage{booktabs}
\usepackage{pifont}
\usepackage{microtype}
\usepackage{inconsolata}
\usepackage{graphicx}
\usepackage{fontawesome5} 
\usepackage{amssymb}
\usepackage{hyperref}
\usepackage{pgfplots}
\usepackage{pgfplotstable}
\usepackage{xcolor}
\usepackage{multirow} 
\pgfplotsset{compat=1.18}
\usepackage{amsmath}

\title{MSA at BEA 2025 Shared Task: Disagreement-Aware Instruction Tuning for Multi-Dimensional Evaluation of LLMs as Math Tutors}

\author{Baraa Hikal, Mohamed Basem, Islam Oshallah, Ali Hamdi \\
  Faculty of Computer Science, MSA University, Egypt \\
  \texttt{\{baraa.moaweya, mohamed.basem1, islam.abdulhakeem, ahamdi\}@msa.edu.eg}
}

\begin{document}
\maketitle

\begin{abstract}
We present \textsc{MSA-MathEval}, our submission to the BEA 2025 Shared Task on evaluating AI tutor responses across four instructional dimensions: Mistake Identification, Mistake Location, Providing Guidance, and Actionability. Our approach uses a unified training pipeline to fine-tune a single instruction-tuned language model across all tracks, without any task-specific architecture modifications. To improve prediction reliability, we introduce a disagreement-aware ensemble inference strategy that enhances coverage of minority labels. Our system achieves strong performance across all tracks, ranking 1\textsuperscript{st} in Providing Guidance, 3\textsuperscript{rd} in Actionability, and 4\textsuperscript{th} in both Mistake Identification and Mistake Location. These results demonstrate the effectiveness of scalable instruction tuning and disagreement-driven modeling for robust, multi-dimensional evaluation of LLMs as educational tutors.
\end{abstract}

\section{Introduction}

Large language models (LLMs) are increasingly used in educational applications, acting as AI tutors that engage students in natural language. However, effective tutoring goes beyond producing correct answers. AI tutors must recognize student mistakes, explain misconceptions, provide constructive guidance, and suggest actionable next steps. Evaluating such complex pedagogical behavior remains a key challenge—especially at scale.

Prior work in intelligent tutoring systems (ITS) emphasized these goals long before the advent of LLMs. For example, AutoTutor used natural language processing (NLP) and dialogue-based feedback to improve learning outcomes across domains \citep{nye-etal-2014-autotutor}. Later, metrics such as \textit{conversational uptake} were proposed to capture tutor responsiveness and its link to instructional quality \citep{demszky-etal-2021-measuring}.

With the rise of instruction-tuned LLMs, evaluation frameworks have emerged to assess their pedagogical abilities. Tack and Piech \citep{tack-piech-2022-teacher} proposed the AI Teacher Test to evaluate whether model responses demonstrate student understanding and helpfulness. Subsequent work introduced fine-grained rubrics, including coherence, correctness, targetedness, and actionability \citep{macina-etal-2023-mathdial, daheim-etal-2024-stepwise, wang-etal-2024-bridging}.

Building on these efforts, the BEA 2025 Shared Task introduces \textit{MRBench}, a pedagogically motivated benchmark for evaluating AI-generated tutor responses in math dialogues \citep{kochmar2025findings}. While BEA 2023 emphasized response generation, BEA 2025 shifts toward assessing feedback quality across four instructional dimensions derived from educational science.

In this work, we present \textsc{MSA-MathEval}, a unified system that addresses all four tracks using a single fine-tuned model and consistent training pipeline. We fine-tune the open-weight \texttt{Mathstral-7B-v0.1}—an instruction-tuned LLM specialized for mathematical reasoning—using parameter-efficient LoRA adapters. To improve prediction reliability, we incorporate ensemble-based inference that combines model disagreement and uncertainty estimation.

\begin{figure*}[t]
    \centering
    \includegraphics[width=\linewidth]{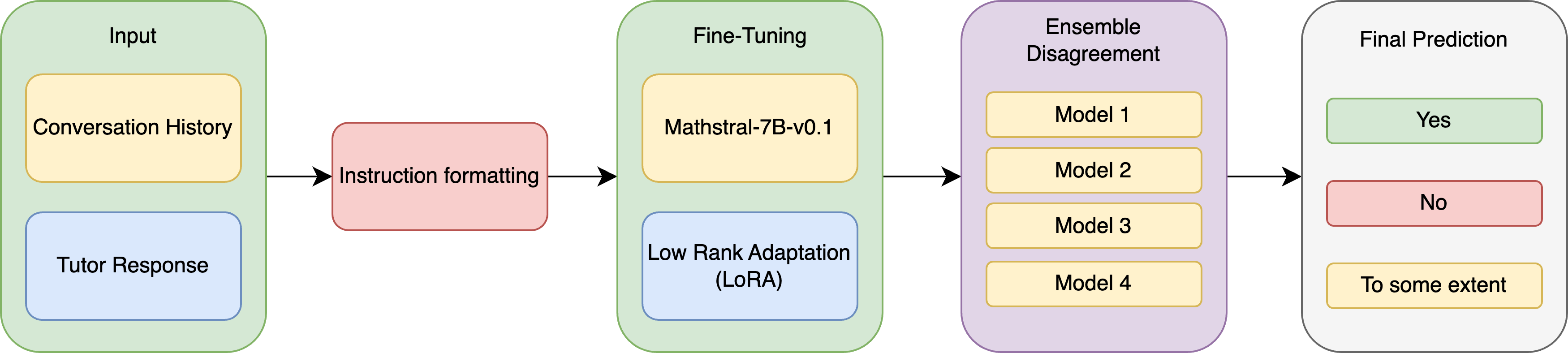}
    \caption{Overview of our unified \textsc{MSA-MathEval} framework for the BEA 2025 Shared Task. The pipeline includes preprocessing, LoRA-based fine-tuning of Mathstral-7B-v0.1, and disagreement-aware ensemble inference.}
    \label{fig:pipeline-overview}
\end{figure*}

\noindent\textbf{Our contributions are as follows:}
\begin{itemize}
    \item We design a unified training pipeline for all four BEA 2025 tracks, using LoRA-based fine-tuning of \texttt{Mathstral-7B-v0.1} with no track-specific architecture changes.
    \item We propose an ensemble-based inference strategy leveraging model disagreement and uncertainty for robust prediction.
    \item We achieve top-tier performance across all tracks, including first place in Providing Guidance.
\end{itemize}

\section{Related Work}

Evaluating the pedagogical capabilities of AI tutors builds upon long-standing research in intelligent tutoring systems (ITS) and more recent advances in large language models (LLMs). Early ITS such as AutoTutor emphasized the importance of natural language dialogue in promoting student learning through error remediation and scaffolding \citep{nye-etal-2014-autotutor}. These systems often relied on rule-based or statistical NLP methods to assess learner inputs and generate appropriate tutor responses.

The emergence of instruction-tuned LLMs has prompted a shift toward more scalable methods for modeling tutoring behavior. Tack and Piech \citep{tack-piech-2022-teacher} proposed the AI Teacher Test to benchmark LLM outputs on criteria such as helpfulness and pedagogical appropriateness. Macina et al. \citep{macina-etal-2023-mathdial} and Daheim et al. \citep{daheim-etal-2024-stepwise} introduced fine-grained rubrics for LLM tutoring quality in mathematical dialogue, including dimensions such as targetedness, coherence, and actionability.

In terms of modeling strategies, prior systems have explored both classification and ranking approaches for feedback generation. Daheim et al. used multi-aspect annotation schemes to evaluate feedback informativeness, while Wang et al. \citep{wang-etal-2024-bridging} proposed a bridging rubric for LLM feedback grounded in human tutor behavior. These studies highlight the need for systems that go beyond correctness to capture richer instructional attributes.

Compared to these approaches, our work introduces a unified training and inference framework across multiple feedback dimensions, leveraging ensemble disagreement and uncertainty estimation for prediction stability. Unlike previous models with track-specific adaptations or rule-based post-processing, we apply a consistent architecture based on the \texttt{Mathstral-7B-v0.1} model across all tasks. This allows us to assess the generalizability of instruction-tuned mathematical LLMs in a multi-dimensional educational evaluation setting.

\section{Method}
Our approach, \textsc{MSA-MathEval}, applies a unified framework across all four tracks in the BEA 2025 Shared Task. We build on the instruction-tuned \texttt{Mathstral-7B-v0.1} model and leverage parameter-efficient fine-tuning (LoRA) along with ensemble-based inference to enhance prediction robustness. The methodology consists of the following stages: dataset preprocessing, model selection, fine-tuning strategy, and ensemble-based inference (see Figure~\ref{fig:pipeline-overview}).

\subsection{Preprocessing}

The original dataset consists of nested JSON files, where each dialogue contains multiple tutor responses annotated across four pedagogical dimensions. To facilitate instruction-based fine-tuning, we transformed the data into four track-specific JSONL files. Each file includes a flattened dialogue, a natural language evaluation prompt, and a categorical label from three possible options: \textit{Yes}, \textit{To some extent}, or \textit{No}.

Each training instance was structured as a two-turn dialogue following the \texttt{chat} schema used by instruction-tuned language models. Specifically:
\begin{itemize}
    \item \textbf{\texttt{user}:} This field contains a complete, track-specific prompt with explicit evaluation criteria, followed by the dialogue context and tutor response to be evaluated.
    \item \textbf{\texttt{assistant}:} This field contains the gold label corresponding to the tutor response—one of \texttt{"Yes"}, \texttt{"To some extent"}, or \texttt{"No"}—as annotated in the development set.
\end{itemize}
The \texttt{system} role was omitted to reduce token overhead and focus the model on the input–output mapping relevant to each multi-class classification task.

\vspace{0.5em}
\hrule
\vspace{0.3em}
\noindent\textbf{Track 1 – Mistake Identification}
\vspace{0.3em}
\hrule

\begin{quote}
\small\ttfamily
TASK DEFINITION:

You are an expert evaluator of AI tutor responses. Your task is to determine whether the tutor's response accurately identifies a mistake in the student's reasoning or solution.

EVALUATION CRITERIA:

1."Yes"– The tutor accurately identifies a mistake in the student’s response.  
2."To some extent"– The tutor shows some awareness, but it is ambiguous or uncertain.\\
3."No"– The tutor fails to identify or misunderstands the mistake.
\end{quote}

\vspace{0.5em}
\hrule
\vspace{0.3em}
\noindent\textbf{Track 2 – Mistake Location}
\vspace{0.3em}
\hrule

\begin{quote}
\small\ttfamily
TASK DEFINITION:

You are an expert evaluator of AI tutor responses. Your task is to determine whether the tutor's response accurately points to a genuine mistake and its location in the student's response.

EVALUATION CRITERIA:

1."Yes"– The tutor clearly points to the exact location of the mistake.  
2."To some extent"– The tutor refers to a mistake but is vague or indirect.  
3."No"– The tutor provides no indication of where the mistake occurred.
\end{quote}

\vspace{0.5em}
\hrule
\vspace{0.3em}
\noindent\textbf{Track 3 – Providing Guidance}
\vspace{0.3em}
\hrule

\begin{quote}
\small\ttfamily
TASK DEFINITION:

You are an expert evaluator of AI tutor responses. Your task is to determine whether the tutor's response provides correct and relevant guidance to help the student.

EVALUATION CRITERIA:

1."Yes"– The tutor gives helpful guidance such as a hint or explanation.\\  
2."To some extent"– The guidance is partially helpful, unclear, or incomplete.\\
3."No"– The guidance is absent, irrelevant, or factually incorrect.
\end{quote}

\vspace{0.5em}
\hrule
\vspace{0.3em}
\noindent\textbf{Track 4 – Actionability}
\vspace{0.3em}
\hrule

\begin{quote}
\small\ttfamily
TASK DEFINITION:

You are an expert evaluator of AI tutor responses. Your task is to determine whether the tutor's feedback is actionable, i.e., it clearly suggests what the student should do next.

EVALUATION CRITERIA:

1."Yes"– The response includes clear next steps for the student.\\ 
2."To some extent"– Some action is implied, but it is not clearly stated.\\  
3."No"– No action is suggested or the feedback ends the conversation.
\end{quote}

\vspace{1em}
Each JSONL instance includes an \texttt{instruction} (as the \texttt{user} message), an \texttt{input} (composed of the full dialogue context and tutor response), and an \texttt{output} (gold label as \texttt{assistant}). This format enables effective supervised fine-tuning of \texttt{Mathstral-7B-v0.1} on each dimension-specific classification task.

\subsection{Model Selection and Architecture}

Our system is built upon the \textit{Mathstral-7B-v0.1} language model, an open-source 7B-parameter Transformer tailored for mathematical and scientific reasoning \citep{mistral2024mathstral}. It is an instruction-tuned variant of the Mistral 7B architecture \citep{jiang2023mistral7b}, which itself builds on the Transformer framework used in LLaMA \citep{touvron2023llama, touvron2023llama2}. Mathstral uses a 32-layer Transformer with 4096-dimensional hidden states and 32 attention heads (8 for keys/values), and benefits from Mistral's sliding-window attention mechanism, enabling long-context comprehension up to 32k tokens. This makes it particularly suitable for modeling multi-turn math tutoring dialogues that require broad context retention.

Mathstral-7B-v0.1 was selected based on its strong performance in math-specific benchmarks and its open-access availability. It was instruction-tuned by Project Numina on mathematical reasoning tasks and achieves high scores on datasets such as GSM8K \citep{cobbe2021gsm8k}, MATH \citep{hendrycks2021math}, and MMLU-STEM \citep{hendrycks2021mmlu}. For instance, it reports 56.6\% accuracy on MATH, significantly outperforming base Mistral and LLaMA models of comparable size.

Compared to alternatives, Mathstral outperforms general-purpose LLaMA 2 \citep{touvron2023llama2} and even surpasses some larger models in mathematical domains. While proprietary models like GPT-3.5 or GPT-4 \citep{openai2022chatgpt, openai2023gpt4} show impressive general capabilities, their closed nature limits fine-tuning flexibility and deployment cost-effectiveness. Mathstral, by contrast, is released under Apache 2.0, making it fine-tunable with LoRA on modest compute budgets.

We thus chose Mathstral-7B-v0.1 as the backbone of our system due to its optimal trade-off between math reasoning accuracy, open weight availability, and instruction-following capability.

\subsection{Training and Fine-Tuning}

We fine-tuned \texttt{Mathstral-7B-v0.1} separately for each BEA 2025 track using Low-Rank Adaptation (LoRA) \citep{hu2021lora}, framing the task as three-way instruction-based classification. Each input was represented as a two-turn dialogue—comprising a prompt and a categorical label—and modeled as a supervised instruction-following task.

\begin{figure}[H]
\centering
\includegraphics[width=0.85\linewidth]{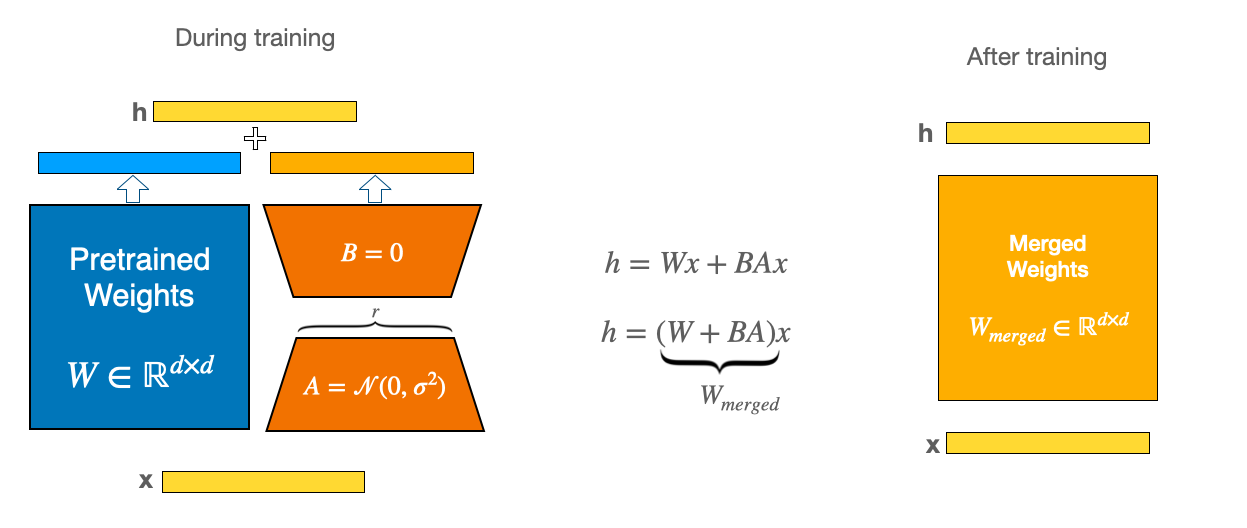}
\caption{LoRA adaptation adds trainable low-rank matrices $A$ and $B$ to frozen attention weights $W_0$, producing an effective weight $W = W_0 + \alpha AB$ during training. Only $A$ and $B$ are updated, enabling memory-efficient fine-tuning \citep{hu2021lora}.}
\label{fig:lora}
\end{figure}

To enable efficient adaptation with minimal memory overhead, we used LoRA with a rank of $r = 64$, scaling factor $\alpha = 2.0$, and no dropout. Adapters were injected into the attention query and value projections in each Transformer block. The low-rank update to the frozen weight matrix $W$ is defined as:
\begin{equation}
\Delta W = \alpha \cdot A B
\end{equation}
where $A \in \mathbb{R}^{d \times r}$ and $B \in \mathbb{R}^{r \times d}$ are trainable matrices, and $d$ is the dimension of the attention head. The final effective weight is $W + \Delta W$. Figure~\ref{fig:lora} illustrates this injection mechanism.

Training was capped at 500 steps with gradient norm clipping ($\|g\|_2 < 1.0$) and a maximum sequence length of 2048 tokens. We used a batch size of 2, single micro-batching, and fixed seed 42 for reproducibility. Optimization was performed using AdamW with a learning rate of $4 \times 10^{-5}$, 10\% linear warmup, and weight decay of 0.05.

We evaluated model performance every 50 steps on a held-out validation set. Checkpoints were saved every 100 steps with a retention window of the three most recent. Only LoRA adapter weights were saved to minimize disk usage and enable efficient inference. All training runs were conducted in a single-node setup with \texttt{world\_size=1}.

This training configuration ensured stable convergence on limited supervision, while maintaining computational efficiency and reproducibility across all four pedagogical dimensions.

\subsection{Inference and Ensemble Strategy}

To enhance robustness and maintain generalization across all four tracks, we employed an ensemble-based inference strategy grounded in model disagreement. Rather than aggregating predictions through majority voting, we fine-tuned five independent models per track and compared their outputs on a per-instance basis. This disagreement-aware mechanism allows us to capture uncertainty and preserve minority-class predictions, especially for ambiguous cases labeled as \texttt{"To some extent"}.

Each model in the ensemble predicts a class independently using greedy decoding. During inference, we collect all five predictions for a given sample and apply a filtering policy: if the predictions exhibit full agreement, the class is retained. If the ensemble disagrees, we analyze the class distribution and prefer predictions that preserve the relative frequency of \texttt{"To some extent"} observed in the development set. This is crucial because \texttt{"Yes"} labels are dominant in both the training and dev sets, potentially leading to biased predictions under a na\"ive voting scheme.

Our design choice is motivated by the use of macro-F1 as the primary evaluation metric in the BEA 2025 Shared Task. Unlike accuracy or micro-F1, macro-F1 gives equal weight to all classes, making performance on minority labels such as \texttt{"To some extent"} especially important. By encouraging the retention of these less frequent but pedagogically relevant labels through disagreement-aware filtering, we improve per-class recall and stabilize final predictions.

\begin{figure*}[t]
    \centering
    \includegraphics[width=\textwidth]{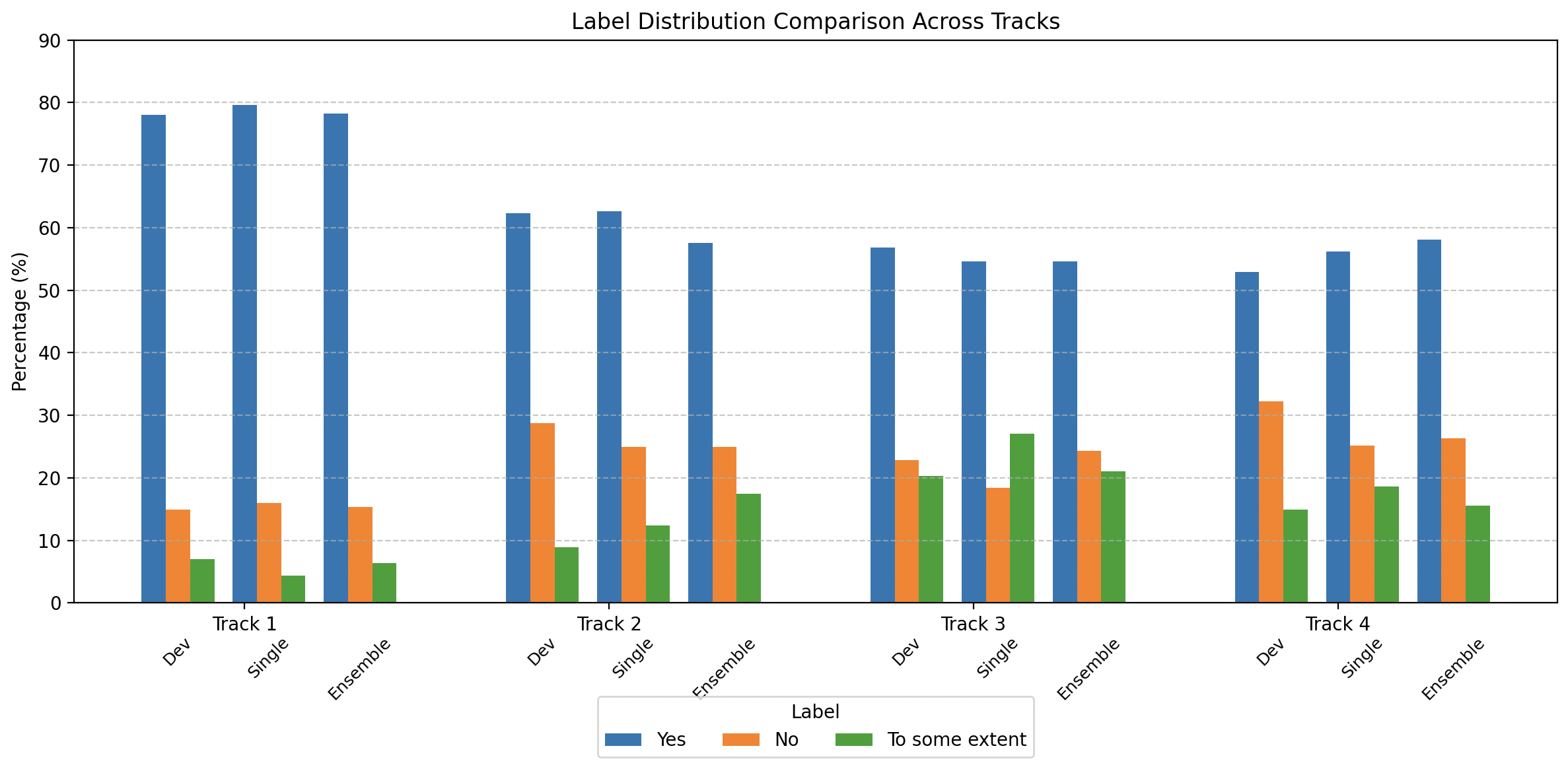}
    \caption{Label distribution comparison across tracks and systems. Each group shows the percentage of predictions per label (\texttt{"Yes"},\texttt{"No"}, \texttt{"To some extent"}, ) for the dev set, single model, and ensemble.}
    \label{fig:label-dist-bar}
\end{figure*}

This ensemble strategy is lightweight in deployment, as only LoRA adapter weights are loaded during inference. Predictions are generated sequentially and combined via a deterministic post-processing script that requires no additional training.

\section{Experiment}
\subsection{Dataset}

The BEA 2025 Shared Task provides a benchmark for evaluating AI tutor responses across four pedagogically motivated tracks: Mistake Identification, Mistake Location, Providing Guidance, and Actionability \citep{kochmar2025findings}. The dataset is based on \textit{MRBench}, a curated collection of math-focused educational dialogues designed for evaluating feedback quality in instructional settings \citep{maurya-etal-2025-unifying}. It includes dialogues drawn from two publicly available sources: MathDial \citep{macina-etal-2023-mathdial} and Bridge \citep{wang-etal-2024-bridging}.

Each instance comprises a multi-turn conversation between a student and an AI tutor, a final student question or statement, and multiple candidate tutor responses. The task is to classify each response along the four instructional dimensions, using a three-way labeling scheme: \textit{Yes}, \textit{To some extent}, and \textit{No}.

The shared task organizers provide a labeled development set with expert annotations for training and validation. The test set is blind—its labels are hidden from participants and used by the organizers to evaluate final system submissions. This setup ensures fair comparison and simulates real-world deployment where labeled data may be limited or unavailable.

\vspace{2.5em}
\noindent\textbf{MRBench Statistics:}
\begin{itemize}
    \item \textbf{192} annotated dialogues in total: \textbf{60} from Bridge and \textbf{132} from MathDial.
    \item \textbf{1,596} total tutor responses annotated across 7 LLMs and multiple human tutors (expert and novice).
    \item Each response is annotated with 8 evaluation dimensions; the shared task focuses on 4 core tracks.
    \item \textbf{Dialogue Length:} Bridge dialogues average 4 turns and 140 characters. MathDial averages 5.5 turns and 906 characters.
\end{itemize}

\subsection{Evaluation}
\label{sec:evaluation}

To evaluate the pedagogical quality of model predictions across all four tracks, the BEA 2025 Shared Task employs two complementary scoring protocols: \textit{exact evaluation} and \textit{lenient evaluation}. Both use macro-averaged F1 score and accuracy as core metrics.

\begin{table*}[t]
\centering
\small
\setlength{\tabcolsep}{6pt}
\begin{tabular}{c|lcccc|c}
\toprule
\textbf{Track} & \textbf{Run} & \textbf{Strict F1} & \textbf{Lenient F1} & \textbf{Strict Acc.} & \textbf{Lenient Acc.} & \textbf{Main Metric Rank} \\
\cmidrule(lr){2-6}

\multirow{5}{*}{\textbf{Mistake Identification}} 
& Run 1 & \textbf{71.54\%} & \textbf{91.52\%} & 87.59\% & \textbf{95.35\%} & \multirow{5}{*}{\textbf{4\textsuperscript{th} / 44}} \\
& Run 2 & 70.66\% & 91.42\% & \textbf{87.98\%} & 95.22\% & \\
& Run 3 & 56.78\% & 82.95\% & 83.65\% & 91.92\% & \\
& Run 4 & 67.88\% & 90.13\% & 87.20\% & 94.76\% & \\
& Run 5 & 71.34\% & \textbf{91.52\%} & 87.39\% & \textbf{95.35\%} & \\
\midrule

\multirow{5}{*}{\textbf{Mistake Location}} 
& Run 1 & 55.62\% & 77.79\% & 72.01\% & 80.93\% & \multirow{5}{*}{\textbf{4\textsuperscript{th} / 31}} \\
& Run 2 & 56.02\% & 77.73\% & \textbf{72.01\%} & 81.19\% & \\
& Run 3 & 56.88\% & \textbf{78.48\%} & 71.88\% & \textbf{82.09\%} & \\
& Run 4 & 52.79\% & 73.65\% & 63.61\% & 78.22\% & \\
& Run 5 & \textbf{57.43\%} & \textbf{78.48\%} & 69.75\% & \textbf{82.09\%} & \\
\midrule

\multirow{4}{*}{\textbf{Providing Guidance}} 
& Run 1 & 55.28\% & 76.02\% & \textbf{67.29\%} & 80.35\% & \multirow{4}{*}{\textbf{1\textsuperscript{st} / 35}} \\
& Run 2 & 53.76\% & \textbf{76.59\%} & 65.09\% & 80.74\% & \\
& Run 3 & 56.65\% & 74.75\% & 63.61\% & 80.61\% & \\
& Run 4 & \textbf{58.33\%} & 77.98\% & 66.13\% & \textbf{81.90\%} & \\
\midrule

\multirow{5}{*}{\textbf{Actionability}} 
& Run 1 & 51.35\% & 68.81\% & 58.31\% & 76.60\% & \multirow{5}{*}{\textbf{3\textsuperscript{rd} / 29}} \\
& Run 2 & 66.99\% & 84.97\% & 71.95\% & 87.91\% & \\
& Run 3 & 65.90\% & 84.45\% & 71.82\% & 87.07\% & \\
& Run 4 & \textbf{69.84\%} & \textbf{86.59\%} & \textbf{75.37\%} & \textbf{89.08\%} & \\
& Run 5 & 65.90\% & 84.45\% & 71.82\% & 87.07\% & \\
\bottomrule
\end{tabular}
\caption{Strict and lenient macro-F1 and accuracy across five runs per track. Bolded scores indicate per-track bests. Final column shows BEA 2025 leaderboard rank based on strict macro-F1 (main metric).}
\label{tab:track-results-detailed}
\end{table*}

\paragraph{Exact Evaluation.}
In the primary setting, each prediction is evaluated against a gold label using a three-way classification scheme: \texttt{"Yes"}, \texttt{"To some extent"}, and \texttt{"No"}. Let $C$ denote the set of all classes, and $F1_c$ the F1 score for class $c \in C$. The macro-F1 score is computed as the unweighted average across all classes:
\begin{equation}
\text{Macro-F1} = \frac{1}{|C|} \sum_{c \in C} \frac{2 \cdot \text{Precision}_c \cdot \text{Recall}_c}{\text{Precision}_c + \text{Recall}_c}
\end{equation}

This metric penalizes class imbalance and rewards systems that maintain recall across minority classes such as \texttt{"To some extent"}.

\paragraph{Lenient Evaluation.}
To account for pedagogical similarity between \texttt{"Yes"} and \texttt{"To some extent"}, the task also includes a two-way lenient evaluation protocol. Labels \texttt{"Yes"} and \texttt{"To some extent"} are merged into a single positive class, resulting in a binary classification task. The same macro-F1 and accuracy metrics are then applied to the collapsed label set.

\paragraph{Accuracy.}
For both settings, accuracy is defined as the proportion of correct predictions over all samples:
\begin{equation}
\text{Accuracy} = \frac{1}{N} \sum_{i=1}^{N} \mathbb{1}(\hat{y}_i = y_i)
\end{equation}
where $N$ is the number of samples, $\hat{y}_i$ is the predicted label, and $y_i$ is the gold label for instance $i$.

\paragraph{Protocol.}
Since the test labels were not released, we computed local metrics only on the development set. All official test results were obtained through the shared task evaluation server. Model selection and early stopping were based on development macro-F1 under the exact evaluation setting, which served as the primary leaderboard metric.

\subsection{Effect of Ensemble Disagreement on Label Distribution}

To analyze the effect of our ensemble strategy on class balance, we examined the label distributions across all four tracks. The development set consistently exhibited a dominant proportion of \texttt{"Yes"} labels—often exceeding 55\%—with \texttt{"To some extent"} and \texttt{"No"} underrepresented.

Left uncorrected, single-model predictions tended to reinforce this imbalance, frequently collapsing uncertain cases into the majority class. To mitigate this, our ensemble disagreement filtering selectively retained predictions for the minority class \texttt{"To some extent"} when model consensus was low. This design choice was informed by the use of macro-F1 as the shared task's official ranking metric, which rewards balanced performance across all classes.
\begin{table*}[t]
\centering
\small
\setlength{\tabcolsep}{8pt}
\begin{tabular}{lcccc}
\toprule
\textbf{Track} & \textbf{Strict Macro-F1} & \textbf{Lenient Macro-F1} & \textbf{Strict Acc.} & \textbf{Lenient Acc.} \\
\midrule
Mistake Identification & \textbf{4\textsuperscript{th}} / 44 & \textbf{2\textsuperscript{nd}} / 44 & \textbf{1\textsuperscript{st}} / 44 & \textbf{2\textsuperscript{nd}} / 44 \\
Mistake Location & \textbf{4\textsuperscript{th}} / 31 & \textbf{6\textsuperscript{th}} / 31 & \textbf{10\textsuperscript{th}} / 31 & \textbf{6\textsuperscript{th}} / 31 \\
Providing Guidance & \textbf{1\textsuperscript{st}} / 35 & \textbf{2\textsuperscript{nd}} / 35 & \textbf{3\textsuperscript{rd}} / 35 & \textbf{3\textsuperscript{rd}} / 35 \\
Actionability & \textbf{3\textsuperscript{rd}} / 29 & \textbf{1\textsuperscript{st}} / 29 & \textbf{2\textsuperscript{nd}} / 29 & \textbf{2\textsuperscript{nd}} / 29 \\
\bottomrule
\end{tabular}
\caption{Per-metric leaderboard ranks (out of all teams) for each track.}
\label{tab:track-leaderboard-ranks}
\end{table*}

Figure~\ref{fig:label-dist-bar} compares label distributions from the development set, single-model outputs, and ensemble predictions. The ensemble strategy improves minority-class coverage—especially for \texttt{"To some extent"}—by better matching the development distribution and mitigating dominant-class bias. This adjustment is particularly useful in ambiguous cases where subtle feedback is warranted.

\section{Results}

We evaluate our system across the four BEA 2025 tracks—Mistake Identification, Mistake Location, Providing Guidance, and Actionability—using both exact (three-class) and lenient (binary) evaluation protocols, as outlined in Section~\ref{sec:evaluation}. We report macro-averaged F1 and accuracy scores across five independent runs for each track and compare our best results to the official leaderboard.

\subsection{Performance Across Runs}

Table~\ref{tab:track-results-detailed} presents detailed performance scores from five independent fine-tuning runs per track. Each run was evaluated on strict and lenient macro-F1 as well as accuracy. We observe moderate variance across runs, particularly in Tracks 2 and 4, which feature more ambiguous tutor responses.

Our best-performing models achieved:
\begin{itemize}
    \item \textbf{Track 1}: 71.54\% strict macro-F1 and 91.52\% lenient macro-F1 (Run 1).
    \item \textbf{Track 2}: 57.43\% strict macro-F1 and 78.48\% lenient macro-F1 (Run 5).
    \item \textbf{Track 3}: 58.33\% strict macro-F1 and 77.98\% lenient macro-F1 (Run 4).
    \item \textbf{Track 4}: 69.84\% strict macro-F1 and 86.59\% lenient macro-F1 (Run 4).
\end{itemize}

These results highlight the robustness of our unified training pipeline and the positive impact of ensemble disagreement filtering on minority-class prediction, especially in borderline cases.

\subsection{Leaderboard Rankings}

Table~\ref{tab:track-leaderboard-ranks} summarizes our official rankings among all participating teams. We consistently placed within the top 5 across all tracks and metrics, securing the \textbf{1\textsuperscript{st}} rank in Track 3 (Providing Guidance) and top-3 ranks in three other metrics.

These ranks validate the effectiveness of our approach across varied pedagogical feedback dimensions. Notably, our system generalizes well across tasks using a unified model and minimal task-specific engineering.

\section{Limitations}

Despite its strong performance across BEA 2025 tracks, our approach has several limitations.

First, the specialization of \texttt{Mathstral-7B-v0.1} to mathematical reasoning may hinder generalization to non-mathematical domains. While domain-specific instruction tuning improves in-domain performance, prior work has shown that such specialization can cause \textit{catastrophic forgetting} of general knowledge, even with parameter-efficient methods like LoRA \citep{dettmers2023qlora}. Moreover, although LoRA significantly reduces memory and compute costs, its low-rank decomposition can constrain the model’s expressiveness in capturing nuanced pedagogical feedback \citep{xu2023parameterefficient, zhou2023revisiting}.

Second, our ensemble disagreement strategy introduces additional inference cost. While it improves recall for minority labels such as \texttt{"To some extent"}, the benefit may diminish if the base models exhibit correlated predictions. Prior work shows that ensembles are most effective when model predictions are diverse and independent \citep{lakshminarayanan2017deep}, which may not always hold in practice.

Finally, the reliance on macro-averaged F$_1$ as the primary evaluation metric, although fair for class imbalance, lacks granularity in penalizing pedagogically critical mistakes. For example, misclassifying a completely wrong tutor response as \texttt{"To some extent"} is penalized equally to a more plausible confusion between \texttt{"Yes"} and \texttt{"To some extent"}. While the lenient evaluation partially addresses this by collapsing similar labels, it does not fully capture the instructional severity of errors \citep{kochmar2025findings}.

\section{Conclusion}

We presented \textsc{MSA-MathEval}, a unified framework for evaluating AI tutor responses across four pedagogical dimensions in the BEA 2025 Shared Task. By fine-tuning a math-specialized LLM (\texttt{Mathstral-7B-v0.1}) using LoRA and leveraging ensemble disagreement during inference, our system achieved top-tier results across all tracks—ranking 1\textsuperscript{st} in Providing Guidance and within the top 5 in all others. Our findings highlight the effectiveness of combining domain-specific instruction tuning with disagreement-aware prediction filtering for educational feedback assessment. Future work will explore cross-domain generalization and dynamic calibration strategies to further enhance robustness.

\bibstyle{acl_natbib}

\end{document}